\pdfoutput=1
\documentclass[conference]{IEEEtran}
\IEEEoverridecommandlockouts
\usepackage{cite}
\usepackage{amsmath,amssymb,amsfonts}
\usepackage{algorithmic}
\usepackage[dvipdfmx]{graphicx}
\usepackage{textcomp}
\usepackage{xcolor}
\usepackage[subrefformat=parens]{subcaption}
\usepackage{bm} 
\usepackage{algorithm,algorithmic}
\usepackage{comment}
\usepackage{color}

\newcommand{\del}[1]{}

\def\BibTeX{{\rm B\kern-.05em{\sc i\kern-.025em b}\kern-.08em
    T\kern-.1667em\lower.7ex\hbox{E}\kern-.125emX}}

\newcommand{\jtextd}[1]{}

\newcommand{\etextd}[1]{}

\begin{document}

\title{
EvolBA: Evolutionary Boundary Attack under Hard-label Black Box condition\\
}

\author{\IEEEauthorblockN{Anonymous Authors}}

\author{\IEEEauthorblockN{Ayane Tajima}
\IEEEauthorblockA{Kagoshima University, Japan} \\
\textit{k3969260@kadai.jp}
\and
\IEEEauthorblockN{Satoshi Ono}
\IEEEauthorblockA{Kagoshima University, Japan} \\
\textit{ono@ibe.kagoshima-u.ac.jp}
}

\maketitle

\begin{abstract}
Research has shown that deep neural networks (DNNs) have vulnerabilities 
that can lead to the misrecognition of Adversarial Examples (AEs) with 
specifically designed perturbations. 
Various adversarial attack methods have been proposed to detect 
vulnerabilities under hard-label black box (HL-BB) conditions 
in the absence of loss gradients and confidence scores.
However, these methods fall into local solutions because
they search only local regions of the search space.
Therefore, this study proposes an adversarial attack method named EvolBA to
generate AEs using Covariance Matrix Adaptation Evolution Strategy
(CMA-ES) under the HL-BB condition, where only a class label predicted by
the target DNN model is available.
Inspired by formula-driven supervised learning, the
proposed method introduces domain-independent operators for
the initialization process and a jump that enhances search
exploration. 
Experimental results confirmed that the proposed method could determine AEs with
smaller perturbations than previous methods in images where the
previous methods have difficulty.

\end{abstract}


\section{Introduction}

Studies over the past decade
have shown that Deep Neural Network (DNN)-based machine
learning models have vulnerabilities that cause erroneous judgments
about Adversarial Examples (AEs) to which special perturbations are
applied~\cite{goodfellow2014explaining}.
Common attack methods to verify the vulnerability of a learner, such as
DNNs, are white-box attacks that generate AEs using the gradient of the
loss function, which is the internal information of the target DNN
models~\cite{goodfellow2014explaining,moosavi2016deepfool,laidlaw2019functional}.
This method is suitable for verifying vulnerabilities in models whose
source code is publicly available.
It is also useful for 
developing commercial models and services to attack their models.
However,  such white-box attack methods cannot be used by a third party to verify
vulnerabilities in proprietary systems or services from the outside.

Therefore, adversarial attack methods under Hard-Label Black Box
(HL-BB) conditions have been widely studied to verify vulnerabilities
without using model internal information
~\cite{brendel2017decision,chen2020hopskipjumpattack,vo2021ramboattack,li2022decision}.
The HL-BB condition is most difficult 
for attackers, where
the information available to the attacker is limited to the label with
the highest confidence.
In other words, adversarial attacks under HL-BB do not use the
internal information of the model, making it possible to verify
commercial models and services whose internal information cannot be
accessed and verify their vulnerabilities by a third party.
HL-BB attacks generally search along the decision boundary while
querying the target model to find AEs with small perturbations,
generating AEs with few perturbations near the decision
boundary without using the confidence scores of the model.

The problems with existing adversarial attacks under the HL-BB condition
are that they search only in the local regions of the
search space, leading to a local solution, and that the search may go
in an inappropriate direction at locations where the objective
function value is fixed, such as a plateau~\cite{vo2021ramboattack}.
Additionally, it is difficult to escape from the local optima
once it has been stuck in them~\cite{vo2021ramboattack}.
The performance of the
attack algorithm also depends on the initial solution (starting
image)~\cite{chen2020hopskipjumpattack,vo2021ramboattack}.
Despite the close relationship between the attack algorithm and the
initial solution, random noise is commonly used as the initial
solution in 
untargeted attacks,
and little research has been
conducted to devise initial solutions.

This study proposes an Evolutionary Boundary Attack (EvolBA),
an adversarial attack method 
using
Covariance Matrix Adaptation Evolution Strategy
(CMA-ES)~\cite{ros2008simple} to generate AEs under the HL-BB condition.
The proposed EvolBA applies three innovations to CMA-ES to search for
a direction that matches the landscape of the objective function:
updating a mean vector 
to avoid the direction of deterioration, 
determining the step size according to the search situation, 
and controlling solutions that deviate from the decision boundary.

Inspired by formula-driven supervised learning (FDSL), the
proposed method introduces domain-independent operators for
initialization and jump operation 
enhance search
exploitation.
Since most convolutional neural networks (CNNs) are sensitive to
high-frequency features~\cite{wang2020high}, adding a high-frequency
pattern formed by mathematical equations as a perturbation enables
good initial solution generation and jumps during the search,
regardless of the characteristics of the target image.
Moreover, it is possible to realize operators that are independent of the 
characteristics of the target image using mathematically generated patterns such as fractals. 

The experimental results showed 
that
EvolBA
found AEs with smaller
perturbations than previous methods in images where the previous 
methods have difficulty~\cite{chen2020hopskipjumpattack}.

The contributions of this study are as follows:
\begin{itemize}

\item {\bf Proposing an evolutionary adversarial attack method under
  the HL-BB condition:}
Although there have been few studies on evolutionary adversarial
attacks under the HL-BB condition, our method achieves
performance that is competitive with the conventional methods.
EvolBA can also incorporate various CMA-ES
improvement techniques 
proposed in the evolutionary
computation community, which will improve the performance of EvolBA
in the future.

\item {\bf Introducing operators inspired by FDSL:}
Recently, FDSL has been 
effectively used
as pre-training
for large-scale DNNs.
This study shows that employing fractal shapes, which is an example of
FDSL, is also effective in adversarial attacks.
In particular, the proposed method does not require domain knowledge of
the target image as well as other DNN models, such as a surrogate
model of the target
DNN~\cite{szegedy2013intriguing,papernot2016limitations}
or an interpreter to obtain a saliency
map~\cite{wu2020boosting}.
\end{itemize}

\section{Related work}

\subsection{Hard-label black-box adversarial attack}

Recently,
HL-BB attacks, which uses only the final
classification result, i.e., the top-1 class label, have been widely
studied because
HL-BB attack can be applicable to commercial systems and services
where internal information such as loss gradient and supplemental
outputs such as confidence scores and top-2 or lower rank labels are
not available~\cite{brendel2017decision,chen2020hopskipjumpattack,maho2021surfree,li2020qeba,li2022decision}.
The HL-BB condition is the most difficult of all the conditions of
adversarial attacks for the following two reasons:
\begin{itemize}

\item Designing an adversarial perturbation for an image is an
  ultra-high-dimensional problem; for instance, in the case of
  attacking a $224\times 224$ pixel image such as the one included in
  ImageNet, it is necessary to determine the values of 150,528
  variables.
  
\item The output from the model of interest is limited to a binary
  value about whether the input perturbation has an affect or not.

\end{itemize}

Brendel et al. proposed a method called Boundary Attack (BA), which
minimizes perturbations under the HL-BB condition by randomly searching
for directions approaching the original image along the boundary
between adversarial and non-adversarial
regions~\cite{brendel2017decision}.
Chen et al. proposed HopSkipJumpAttack (HSJA), that estimates the
gradient of the decision boundary by approximating the local decision
boundary using the Monte Carlo method~\cite{chen2020hopskipjumpattack}.
Instead of a fine approximation of the decision boundary, HSJA
presents a new gradient estimation method that minimizes the amount of
perturbation by estimating the direction
to the decision boundary surface from the region near the AE and
combining it with a binary search.
This method has a high ability to estimate the gradient of the
decision boundary and minimizes the perturbation with a small number
of queries to the target model.
However, it may fall into the local optima because it only refers to
local decision boundaries during the search, and it may also
incorrectly estimate the gradient at plateaus and local minima.

Li et al. proposed f-mixup for generating AEs by focusing on
high-frequency components in the frequency domain and a frequency
binary search based on f-mixup\cite{li2022decision}.
Based on the fact that the existence of AEs can be attributed to the
corruption of useful features and the
introduction of misleading information,
they proposed the method to replace high frequency components of an
original image with that of another natural image, allowing them to
generate AEs with less noticeable perturbations while
preserving the semantic features of the original image.

On the other hand, it has been pointed out that the performance of
attack algorithms depends on the initial solution in adversarial
attack methods under HL-BB conditions, and that some initial solutions
can lead the algorithm to local solutions of low
quality~\cite{chen2020hopskipjumpattack,vo2021ramboattack}.
As a device for initial solutions, perturbation obtained using
transition attacks can be used as initial solutions to improve the
performance of adversary attack methods under the HL-BB
conditions~\cite{moosavi2017universal}.
This method used past attack cases in the target domain as training
data, and to the best of the authors' knowledge, an efficient method
for generating initial solutions that does not require prior knowledge
has not yet been proposed.
It is also known that it is difficult to escape from a local solution
once it is stuck in a local solution, even if a enough number of
queries are spent on it.

\subsection{Formula-driven supervised learning}

Kataoka et al. proposed a method to generate image patterns and their
category labels by assigning fractal values based on natural laws
existing in real-world background knowledge instead of natural
images~\cite{kataoka2020pre}.
This method employs fractals generated by simple mathematical
formulas\cite{MAN83}, which are closely related to natural objects
composed of complex patterns.
It was shown that a model that is pre-trained on their fractal dataset,
which does not include any natural images, partially exceeded the
accuracy of a model pretrained on ImageNet.

\begin{figure}[t]
  {\scriptsize
  \begin{minipage}[t]{0.48\hsize}
     \centering
     \includegraphics[width=4.0cm]{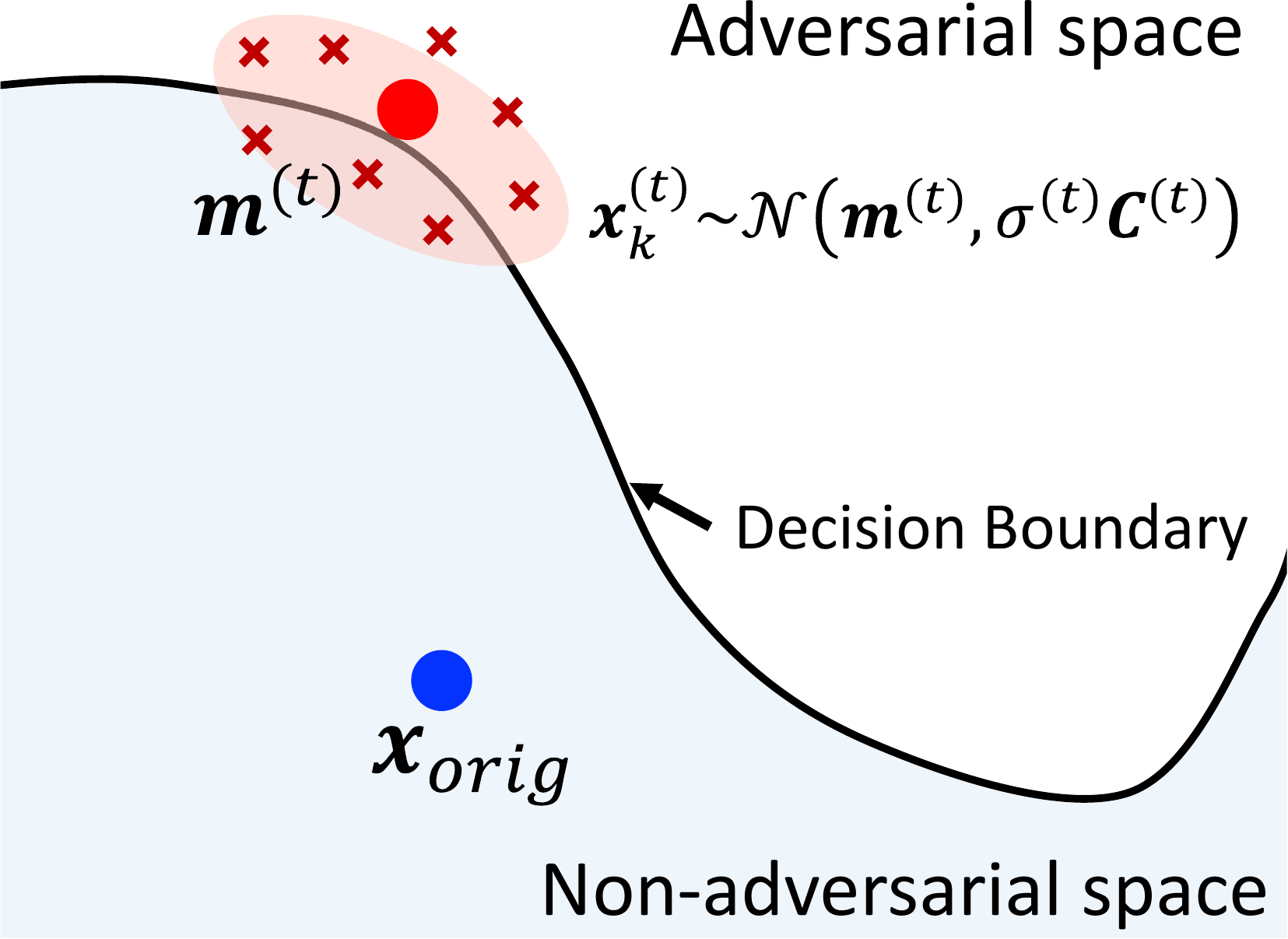}
      \subcaption{Sampling and evaluating offspring}
  \end{minipage}
  \begin{minipage}[t]{0.48\hsize}
     \centering
     \includegraphics[width=4.0cm]{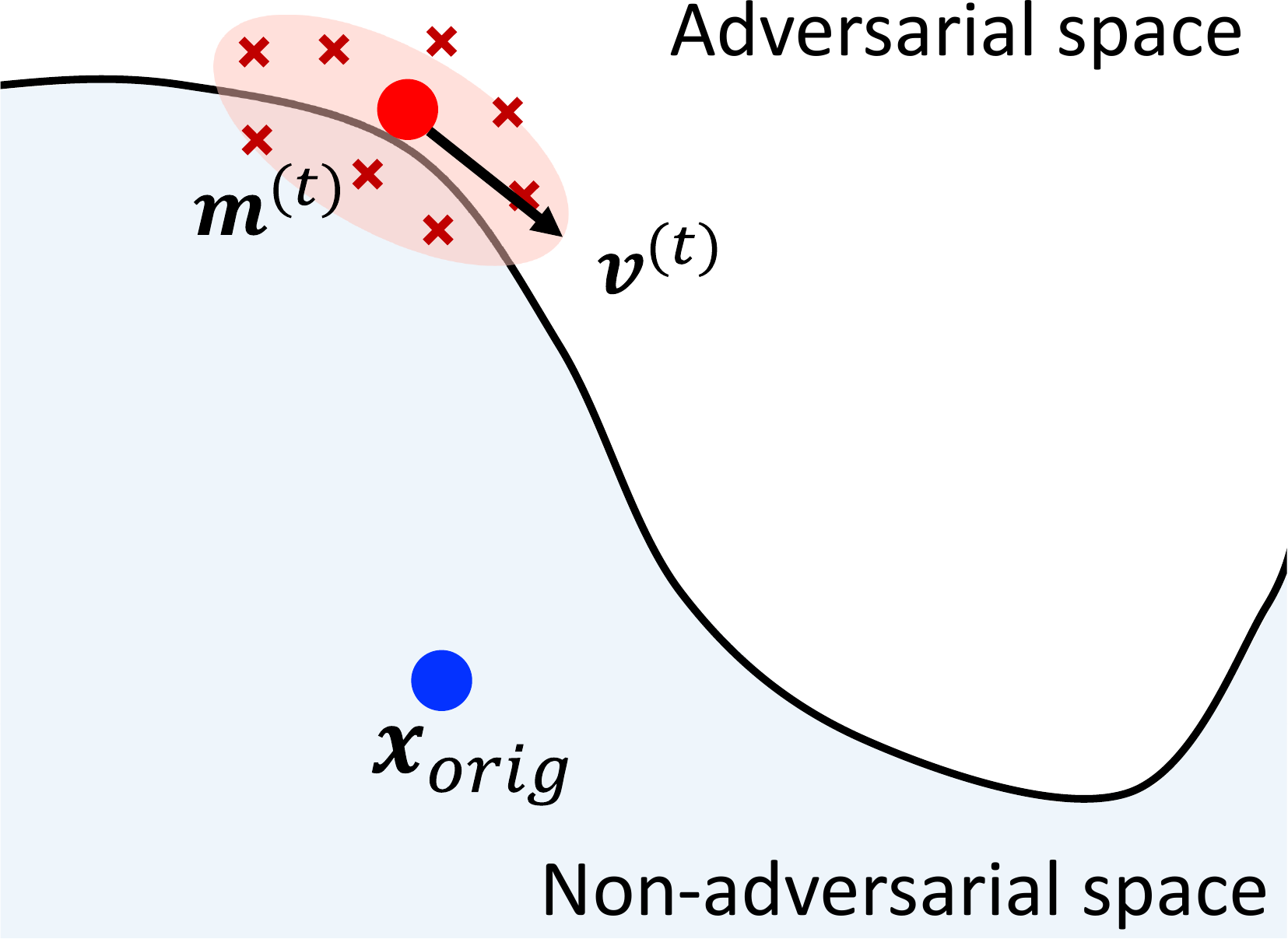}
     \subcaption{Moving on the decision boundary}
  \end{minipage}
  \\
  \begin{minipage}[t]{0.48\hsize}
    \centering
    \includegraphics[width=4.0cm]{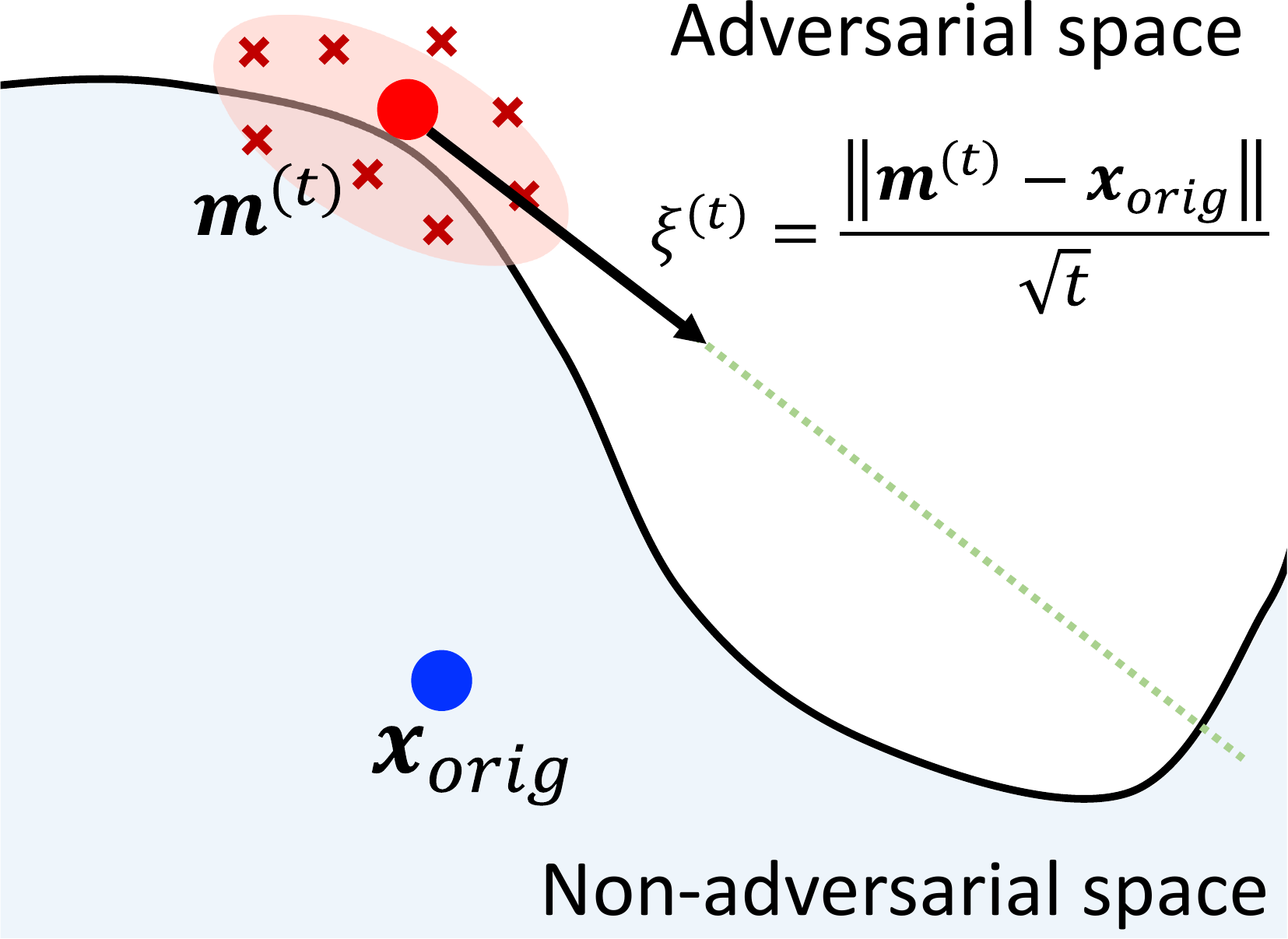}
    \subcaption{Step-size adaptation}
 \end{minipage}
 \begin{minipage}[t]{0.48\hsize}
    \centering
      \includegraphics[width=4.0cm]{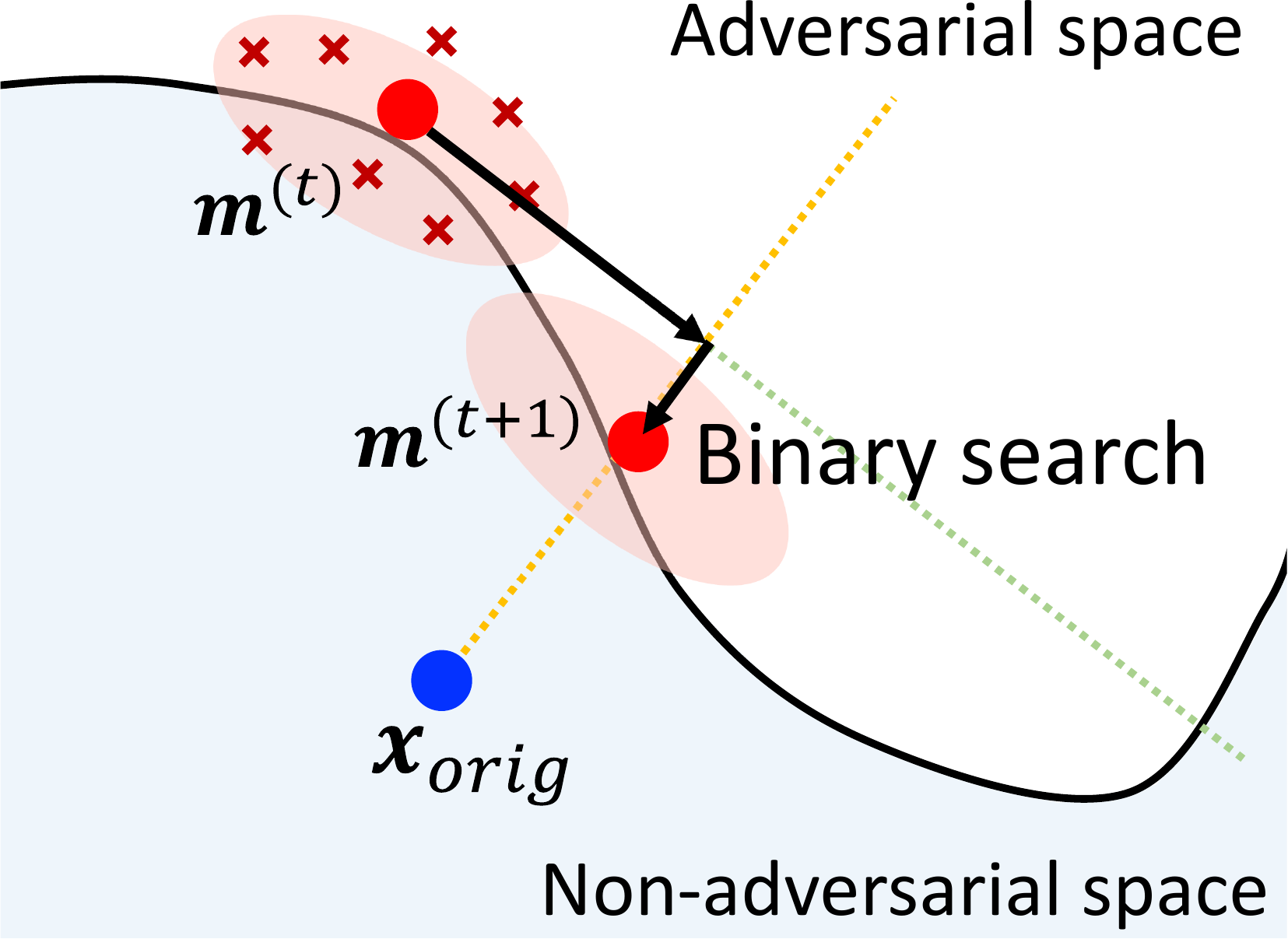}
    \subcaption{Moving towrad the original image}
 \end{minipage}
}
  \caption{Overview of the proposed EvolBA.}
  \label{fig:CMAES03}
\end{figure}
\section{Proposed evolutionary boundary attack method (EvolBA)}

\subsection{Key ideas}

The proposed EvolBA is an evolutionary adversarial attack method that
discovers AEs of the target DNN model under the HL-BB
condition.
The key ideas of EvolBA
are summarized as follows:

\hspace*{-3ex}
{\bf Idea 1: Hybridizing Sep-CMA-ES~\cite{ros2008simple} and Boundary Attack (BA)~\cite{brendel2017decision}.}
To solve extremely high-dimensional optimization problems, the
proposed method employs Sep-CMA-ES, which reduces spatial and temporal
computational complexity by restricting a covariance matrix to its
diagonal components.
When searching on the decision boundary, 
EvolBA
updates the
mean vector of Sep-CMA-ES in a way specific to the 
AE
generation problem and adaptively changes the step size according to
the search situation to find the direction that matches the landscape
shape of the objective function.

\hspace*{-3ex}
{\bf Idea 2: Initial solution generation using fractal image features.}
Focusing on the CNNs' characteristics, 
which are sensitive to
high-frequency features, the proposed method generates an initial
solution by combining the high-frequency components of a fractal
image~\cite{kataoka2020pre} with a given original image using discrete
Fourier transform (DFT).
This method 
can
efficiently provide an initial AE candidate without requiring domain
knowledge or prior learning
since it uses only the original and fractal images.

\hspace*{-3ex}
{\bf Idea 3: A jump operator using fractal image features.}
Similar to BA, 
EvolBA
generates AEs by repeatedly moving
on the discriminant boundary and
toward the
original image, 
thereby enabling efficient exploration of
ultra-high-dimensional space. 
On the other hand, 
the limited search area makes search exploration
weak.
Therefore, as in Idea 2, the proposed method introduces a jump
operator that uses fractal image features.
Moreover, EvolBA can change the search area by convolving the high-frequency component of the fractal image with
the AE candidate.

The operators for initial solution generation and jump of the proposed method 
were inspired by FDSL~\cite{kataoka2020pre}, 
in which a pre-training dataset for image
classification was created from fractal patterns based on natural laws.
When generating AEs, superimposing fractal features on the original
image weakens the features of the original image and adds new
features, 
thereby effectively inducing misrecognition of the
target DNN.
Additionally, these operators merge fractal features in frequency
space using DFT, 
similar to Li's method~\cite{li2022decision}, which showed
that frequency binary search significantly changed the search region
compared to a binary search in the spatial domain.

Note that the jump operator proposed in this paper differs from
that of HSJA, although they have the same name.

\subsection{Formulation}

\subsubsection{Variables}

Similar to general methods of adversarial attacks in image
recognition, 
EvolBA
directly optimizes perturbations to
all pixels in all color channels of a target image.
That is, the total number of design variables $N$ becomes $3wh$ where
$w \times h$ denotes a resolution of the original image $\bm{x}$ and the
number of color channels is $3$.
This is an extremely high-dimensional optimization problem for
algorithms that do not use the gradient of the objective function;
however, Sep-CMA-ES in 
EvolBA
allows direct optimization
without reducing the dimension.

\subsubsection{Objective function}

Under
SL-BB condition, 
it is common to minimize the confidence score of the correct class 
the target DNN outputs to find an AE.
In contrast, under the HL-BB condition where only the Top-1 class labels
are available and confidence levels cannot be referenced, optimization
must be performed to minimize the amount of perturbation.
Therefore, the objective function of 
EvolBA
is the L2
norm of the perturbation plus a penalty function:
\begin{eqnarray}
   {\rm minimize} && \quad  f(\bm{\tilde{x}}) = ||\bm{\tilde{x}} - \bm{x}_{\rm orig} || + f_p(\bm{\tilde{x}})  \\
   {\rm subject~to} && \quad  \mathcal{C}(\bm{\tilde{x}})\neq \mathcal{C}( \bm{x}_{orig} ) 
\end{eqnarray}
where
$\bm{\tilde{x}}$ and $\bm{x}_{\rm orig} $ denote
AE candidate and original image, respectively.
$f_p$ is a penalty function that is applied to non-adversarial
offspring to clearly distinguish between adversarial and
non-adversarial offspring.
$\mathcal{C}(\cdot)$ denotes the Top-1 class label to $\bm{\tilde{x}}$
assigned by the target classifier.
\begin{align}\label{eq:penalty}
    f_{p}(\bm{\tilde{x}}) = \left\{
    \begin{array}{ll}
        c_{pen} & \mbox{if~ $\mathcal{C}(\bm{\tilde{x}})= \mathcal{C}( \bm{x}_{orig} )$} \\
        0 & \mbox{otherwise}
    \end{array}
    \right.
\end{align}

\subsection{Algorithm of EvolBA}

EvolBA
is based on the hybridization of
Sep-CMA-ES~\cite{ros2008simple} and Boundary Attack
(BA)~\cite{brendel2017decision}.
EvolBA
optimizes an AE by repeating four main processes:
(a) sampling and evaluating offsipring,
(b) moving on the decision boundary
(c) step-size adaptation, and 
(d) moving toward the original image,
as shown in Fig.~\ref{fig:CMAES03}.
It applies Sep-CMA-ES to the search on the decision boundary, which is
the discriminative boundary between classes obtained by a target DNN,
in the feature space.

Since HSJA~\cite{chen2020hopskipjumpattack}, one of the previous
studies, is also based on BA, the algorithm of EvolBA is
similar to the HSJA algorithm but there are differences such as the
movement along the boundary and the operators using fractal images.

  \begin{algorithm}[t]
      \caption{EvolBA}
      \label{alg:proposedMethod}
      {\small
        \begin{algorithmic}[1]
            \STATE $t \leftarrow 0$
            \STATE Create initial AE candidate $\bm{x}^{(0)}$ by
            a uniform random noise  
            \STATE Apply binary search between $\bm{x}_{orig}$ and
            $\bm{x}^{(0)}$ to move to the decision boundary
            \STATE $\tilde{\bm{x}}^{(0)} \leftarrow \text{Bin-Search}(\bm{x}^{(0)},\bm{x}_{orig})$
            \STATE Initialize Sep-CMA-ES parameters and  $\bm{m}^{(0)} \leftarrow \tilde{\bm{x}}^{(0)}$
            \WHILE{The number of queries reaches the limit}
                \STATE Sample offspring
                $\bm{x}^{(t)}_{k}(k=1,2,...,\lambda)$ from
                $\mathcal{N}(\bm{m}^{(t)},
                       {\sigma^{(t)}}^{2}\bm{C}^{(t)})$
                \STATE Evaluate $\bm{x}^{(t)}_{k}(k=1,2,...,\lambda)$
                \STATE Update Sep-CMA-ES parameters, $\bm{v}^{(t)}$ and $\bm{C}^{(t+1)}$
                \STATE // Determine step size $\xi^{(t)}$
                \STATE $\xi^{(t)}\leftarrow||\bm{m}^{(t)} - \bm{x}_{orig}|| / \sqrt{t}$
                \WHILE {$\mathcal{C}(\bm{m}^{(t)}+\xi^{(t)}\bm{v}^{(t)}) = \mathcal{C}(\bm{x}_{orig})$}
                    \STATE $\xi^{(t)} \leftarrow \xi^{(t)}/2$
                \ENDWHILE
                \STATE // Move $\bm{m}^{(t+1)}$ along the decision boundary
                \STATE $\bm{m}^{(t+1)} \leftarrow\bm{m}^{(t)}+\xi^{(t)}\bm{v}^{(t)}$
                \STATE // Move $\bm{m}^{(t+1)}$ toward $\bm{x}_{orig}$ 
                \STATE $\bm{m}^{(t+1)} \leftarrow \text{Bin-Search}(\bm{m}^{(t+1)},\bm{x}_{orig})$
                \WHILE{$||\bm{m}^{(t)} - \bm{x}_{orig}|| < ||\bm{m}^{(t+1)}-\bm{x}_{orig}||$} 
                    \STATE $\xi^{(t)} \leftarrow \xi^{(t)}/2$ 
                    \STATE $\bm{m}^{(t+1)} \leftarrow \text{Bin-Search}(\bm{m}^{(t)} + \xi^{(t)}\bm{v}^{(t)},\bm{x}_{orig})$
                \ENDWHILE
                \STATE $t \leftarrow t+1$
          \ENDWHILE
        \end{algorithmic}
      }
  \end{algorithm}

  \begin{algorithm}[t]
      \caption{Binary search}
      \label{alg:Bin-Search}
      {\small
        \begin{algorithmic}[1]
          \STATE Let adversaial and non-adversarial points be $\bm{x}'$ and $\bm{x}$
          \STATE $\alpha_{l} \leftarrow 0$, $\alpha_{u} \leftarrow 1$
          \WHILE{Not reached the iteration limit}
            \STATE $\alpha_{m} \leftarrow \frac{\alpha_{l}+\alpha_{u}}{2}$
            \IF {$\prod_{\bm{x},\alpha_{m}}(\bm{x}')$ is adversarial}
                \STATE $\alpha_{u} \leftarrow \alpha_{m}$
            \ELSE 
                \STATE $\alpha_{l} \leftarrow \alpha_{m}$
            \ENDIF
          \ENDWHILE
          \STATE Return $\prod_{\bm{x},\alpha_{u}}(\bm{x}')$
        \end{algorithmic}
      }
  \end{algorithm}

  \begin{algorithm}[t]
      \caption{Initialization}
      \label{alg:Initialization}
      {\small
        \begin{algorithmic}[1]
          \REPEAT
              \STATE $\bm{f}_o \leftarrow \text{DFT}(\bm{x}_{\rm orig})$, $\bm{f}_i \leftarrow \text{DFT}(\bm{x}_i)$
              \STATE $\bm{f}_o^{lp} \leftarrow \text{lp}(\bm{f}_o; r)$, $\bm{f}_i^{hp} \leftarrow \text{hp}(\bm{f}_i; r)$
              \STATE $\bm{f}_{oi} \leftarrow \bm{f}_o^{lp} +\bm{f}_i^{hp}$
              \STATE $\bm{x}_{oi} \leftarrow \text{IDFT}(\bm{f}_{oi})$
              \STATE $r \leftarrow r - 1$
          \UNTIL{$\mathcal{C}(\bm{x}_{oi}) \neq \mathcal{C}( \bm{x}_{\rm orig} )$}
          \STATE Return a generated AE candidate $\bm{x}_{oi}$
        \end{algorithmic}
      }
  \end{algorithm}
\begin{figure}[t]
  \centering
  \includegraphics[width=1.0\linewidth]{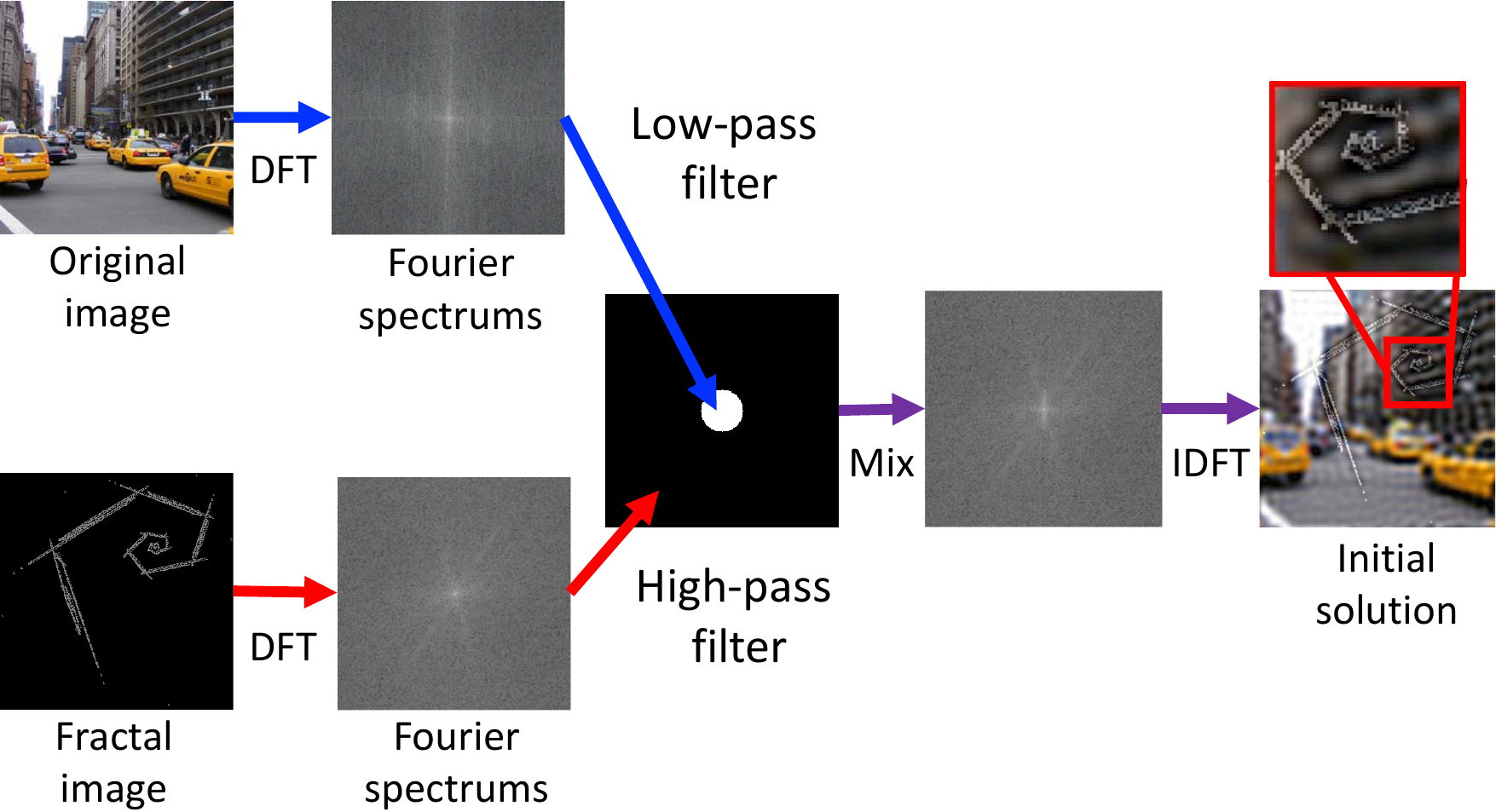}
  \caption{Initial solution generation processes.}
  \label{fig:Initialization}
\end{figure}
\begin{figure}[t]
  {\scriptsize
  \begin{minipage}[b]{0.32\hsize}
    \centering
    \includegraphics[width=2.5cm]{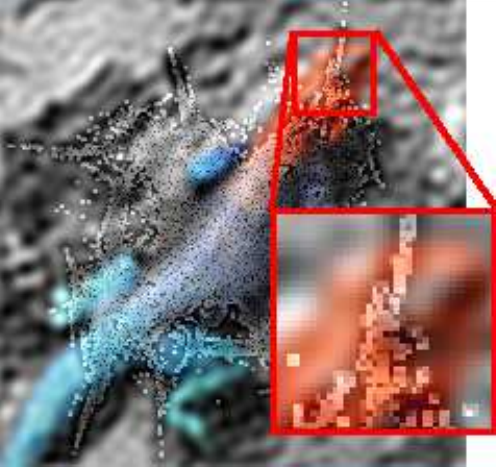}
    \subcaption{Initial solution example.}
  \end{minipage} 
  \begin{minipage}[b]{0.32\hsize}
    \centering
    \includegraphics[width=2.5cm]{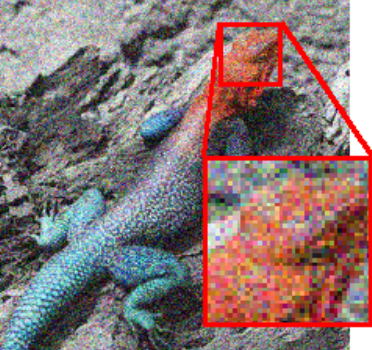}
    \subcaption{AE candidate before jumping.}
  \end{minipage} 
  \begin{minipage}[b]{0.32\hsize}
    \centering
    \includegraphics[width=2.5cm]{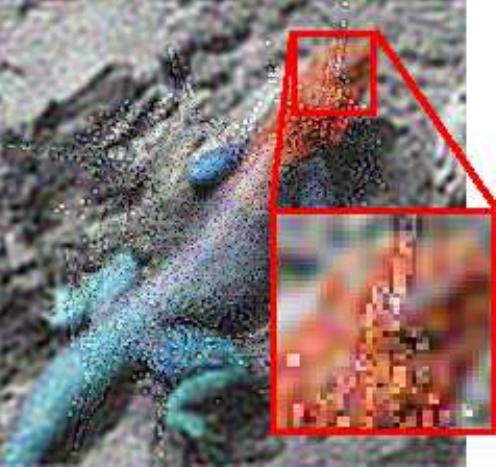}
    \subcaption{AE candidate after jumping.}
  \end{minipage}
  }
  \caption{Examples of initialization and jump operator application.}
  \label{fig:AE_I+J}
\end{figure}

The detailed algorithm of EvolBA
is shown in
Algorithm~\ref{alg:proposedMethod} and below.
In the initialization process (l. 1-5),
EvolBA creates an initial AE candidate $\bm{x}^{(0)}$ by 
a uniform random noise  
and moving it on the
decision boundary by binary search, and 
let $\bm{x}^{(0)}$ be an initial mean vector of a multivariate
normal distribution of Sep-CMA-ES.

In the main loop, EvolBA 
samples $\lambda=4+ \lfloor 3 \text{ln}(N)\rfloor$ offspring
according to the distribution (l. 7).
Then, it applies a target DNN model to the sampled offspring
and calculate their objective
function values (l. 8).

When moving on the decision boundary,
EvolBA
determines a step size $\xi^{(t)}$
as follows:
\begin{equation}
\xi^{(t)}=\frac{||\bm{m}^{(t)} - \bm{x}_{orig}||}{\sqrt{t}}
\end{equation}
If the updated mean vector is in a non-adversarial region, 
repeat the operation of halving the step size 
until the vector will move to an adversarial region (l. 10-14).

After the mean vector moved on the decision boundary surface,
it
moves toward
the original image by applying binary search whose algorithm is described in
Algorithm~\ref{alg:Bin-Search}.
However, if the direction of movement on the boundary surface
is not appropriate, or if the
objective function value after binary search is worse than the current
value, halve the step size and go back to l.~16, i.e., calculate
$\bm{m}^{(t+1)}$ again.
The number of times to the backtrack is limited to $\tau$%
\footnote{%
To simplify the pseudo code, the description of the upper limit of the
number of iterations is omitted in Algorithm\ref{alg:proposedMethod}.
}.
That is, if the objective function value does not improve after
$\tau$
times of reducing step size and applying binary search, then do not
update the mean vector and proceed to the next generation.

\subsection{Initialization using a fractal image}
\label{ssec:gen_init_sol}

Algorithm~\ref{alg:Initialization} and Figs.~\ref{fig:Initialization} show the proposed initial
solution generation algorithm and its scheme.
In order to make the solution candidate $\bm{x}_{oi}$ obtained using the
frequency domain, the proposed method repeats the following processes
while lowering the cutoff frequency $r$ to reduce the low frequency
component of the original image.
First, we use DFT to convert two images, an original image $\bm{x}_{\rm orig}$ and a
fractal image $\bm{x_i}$, from the spatial domain to the frequency domain.
\begin{eqnarray}\label{eq:}
  \bm{f}_o = \text{DFT}(\bm{x}_{\rm orig}), \quad && \bm{f}_i = \text{DFT}(\bm{x}_i)
\end{eqnarray}
Then, the two images are combined by replacing the high-frequency components of the original image
with those of the fractal image:
\begin{eqnarray}\label{eq:}
  \bm{f}_{oi} = \bm{f}_o^{lp} +\bm{f}_i^{hp}
\end{eqnarray}
\begin{eqnarray}\label{eq:}
  \bm{f}_o^{lp} = \text{lp}(\bm{f}_o; r), \quad && \bm{f}_i^{hp} = \text{hp}(\bm{f}_i; r)
\end{eqnarray}
where $\text{lp}(\cdot; \cdot)$ and $\rm{hp}(\cdot; \cdot)$ denote low
pass and high pass filters with cut-off frequency $r$, respectively.
Finally, an inverse discrete Fourier transform (IDFT) is used to
transform from the frequency domain to the spatial domain, and the
initial solution image is obtained from the combined frequency
components.
\begin{eqnarray}\label{eq:}
  \bm{x}_{oi} = \text{IDFT}(\bm{f}_{oi})
\end{eqnarray}
The resulting initial AE candidate is an image in which the fractal
image edges are embedded on the smoothed original image, as shown in
Fig.~\ref{fig:AE_I+J}(a).

\subsection{Jump operator}

CMA-ES is a relatively local optimization algorithm compared to other
evolutionary algorithms, and early convergence to a local solution can
be problematic.
To enhance search exploration, the proposed method introduces a jump
operator using a fractal image as the initialization operator
described in Sec~\ref{ssec:gen_init_sol}.
Algorithm~\ref{alg:restart} shows algorithm of the jump operator.
The point at which the jump is made is calculated in the same manner
as the initialization, i.e., merging the high-frequency components of the
fractal image $\bm{x}_i$ to mean vector $\bm{m}^{(t)}$ with cutoff
frequency $r$.
Fig.~\ref{fig:AE_I+J}(b) and (c) show an example AE candidate change
by the jump operator, which shows that the perturbation being
optimized and fractal features are fused by the jump.

  \begin{algorithm}[t]
      \caption{Jump operator}
      \label{alg:restart}
      {\small
        \begin{algorithmic}[1]
              \STATE $\bm{f}_t \leftarrow \text{DFT}(\bm{m}^{(t)})$, $\bm{f}_i \leftarrow \text{DFT}(\bm{x}_i)$
              \STATE $\bm{f}_t^{lp} \leftarrow \text{lp}(\bm{f}_t; r)$, $\bm{f}_i^{hp} \leftarrow \text{hp}(\bm{f}_i; r)$
              \STATE $\bm{f}_{ti} \leftarrow \bm{f}_t^{lp} +\bm{f}_i^{hp}$
              \STATE $\bm{m}^{(t)} \leftarrow \text{IDFT}(\bm{f}_{ti})$
          \STATE Return jumped mean vector $\bm{m}^{(t)}$
        \end{algorithmic}
      }
  \end{algorithm}

\subsection{Update a mean vector of Sep-CMA-ES}
\label{ssec:update_mean}

In the problem of generating AEs under the HL-BB condition, a general mean
vector update calculation of Sep-CMA-ES causes the mean vector to move
toward the original image and invade the non-adversarial region,
making it difficult to generate AE.
For this reason,  
EvolBA
uses the offspring generated in
adversarial to update the mean vector,
thereby avoiding
the mean vector from intruding into the non-adversarial region.
Furthermore, the sampled offspring are weighted in the order
of their evaluation values to avoid moving in the direction of
increasing perturbations, i.e., away from the original
image. 
Individuals that have entered the non-adversarial region are placed
in the adversarial region by multiplying by $-1$. 

The following equation updates the mean vector of the
distribution in 
EvolBA
based on the above policy.
\begin{align}\label{eq:mean_AE}
  \bm{m}^{(t+1)} = \bm{m}^{(t)} +  \xi^{(t)} \bm{v}^{(t)}
\end{align}
\begin{align}\label{eq:cov_sep}
  \bm{v}^{(t)}=\frac{\sum_{i=1}^{l}  w_i\bm{z}^{(t)}_{i:\mu}-\sum_{j=l+1}^{\mu} w_{j}\bm{z}^{(t)}_{(\mu-j+l+1):\mu}}{||\sum_{i=1}^{l} w_i\bm{z}^{(t)}_{i:\mu}-\sum_{j=l+1}^{\mu} w_j\bm{z}^{(t)}_{(\mu-j+l+1):\mu}||}
\end{align}
where $\bm{z}_{i:\mu}$ corresponds to the $i$-th best-valued adversarial
solution candidate and 
$\bm{z}_{(\mu-j+l+1):\mu}$
corresponds to the $j$-th
worst-valued non-adversarial solution candidate.
The total number of adversarial solution candidates is $l$ ($ \leq
\mu$) and the number of non-adversarial solution candidates is $\mu -
l$.

Since it is necessary to search in an extremely high-dimensional space
and even non-adversarial offspring can provide clues,
EvolBA
employs all offspring to update the mean vector,
i.e., $\mu = \lambda$, while usual Sep-CMA-ES uses only good
individuals, e.g, $\mu = \lfloor \lambda / 2 \rfloor$.

\begin{table}[t]
  \centering
  \caption{Pre-Exp 1: the effect of $\sigma$}
  \begin{tabular}{c c c|r} 
      \hline
      $c_{\mu}$&$c_{1}$&$\sigma$& $l_2$ distance\\
      \hline \hline
      $8.89 \times 10^{-5}$&$4.42 \times 10^{-6}$&$1.0$&60.51\\
      $8.89 \times 10^{-5}$&$4.42 \times 10^{-6}$&$1.0 \times 10^{-1}$&19.21\\
      $8.89 \times 10^{-5}$&$4.42 \times 10^{-6}$&$1.0 \times 10^{-2}$&5.28\\
      $8.89 \times 10^{-5}$&$4.42 \times 10^{-6}$&$\bm{1.0} \times \bm{10^{-3}}$&\textbf{2.37}\\
      $8.89 \times 10^{-5}$&$4.42 \times 10^{-6}$&$1.0 \times 10^{-4}$&2.52\\
      $8.89 \times 10^{-5}$&$4.42 \times 10^{-6}$&$1.0 \times 10^{-5}$&2.49\\
      $8.89 \times 10^{-5}$&$4.42 \times 10^{-6}$&$1.0 \times 10^{-6}$&2.62\\
      \hline
  \end{tabular}
\label{tab:Experiment_sigma}
~\\ ~\\
  \caption{Pre-Exp 2: the effect of $c_{\mu}$}
  \begin{tabular}{c c c|r} 
      \hline
      $c_{\mu}$&$c_{1}$&$\sigma$& $l_2$ distance\\
      \hline \hline
      $8.89 \times 10^{-5}$&$4.42 \times 10^{-6}$&$1.00 \times 10^{-3}$&2.37\\
      $1.00 \times 10^{-5}$&$4.42 \times 10^{-6}$&$1.00 \times 10^{-3}$&2.39\\
      $1.00 \times 10^{-4}$&$4.42 \times 10^{-6}$&$1.00 \times 10^{-3}$&2.35\\
      $1.00 \times 10^{-3}$&$4.42 \times 10^{-6}$&$1.00 \times 10^{-3}$&2.48\\
      $1.00 \times 10^{-2}$&$4.42 \times 10^{-6}$&$1.00 \times 10^{-3}$&2.38\\
      $\bm{1.00} \times \bm{10^{-1}}$&$4.42 \times 10^{-6}$&$1.00 \times 10^{-3}$&\textbf{2.19}\\
      $2.00 \times 10^{-1}$&$4.42 \times 10^{-6}$&$1.00 \times 10^{-3}$&2.27\\
      $3.00 \times 10^{-1}$&$4.42 \times 10^{-6}$&$1.00 \times 10^{-3}$&2.23\\
      $4.00 \times 10^{-1}$&$4.42 \times 10^{-6}$&$1.00 \times 10^{-3}$&2.43\\
      \hline
    \end{tabular}
  \label{tab:Experiment_cmu}
~\\ ~\\
  \caption{Pre-Exp 3: the effect of $\lambda$}
    \begin{tabular}{c c c|r} 
        \hline
        $c_{\mu}$&$\sigma$&$\lambda$& $l_2$ distance\\
        \hline \hline
        $1.00 \times 10^{-1}$&$1.00 \times 10^{-3}$&39(defoult)&2.26\\
        $1.00 \times 10^{-1}$&$1.00 \times 10^{-3}$&78&2.19\\
        $1.00 \times 10^{-1}$&$1.00 \times 10^{-3}$&\textbf{117}&\textbf{1.96}\\
        $1.00 \times 10^{-1}$&$1.00 \times 10^{-3}$&156&2.17\\
        $1.00 \times 10^{-1}$&$1.00 \times 10^{-3}$&195&2.02\\
        $1.00 \times 10^{-1}$&$1.00 \times 10^{-3}$&234&2.09\\
        \hline
  \end{tabular}
\label{tab:Experiment_lambda}
~\\ ~\\~\\
  \caption{Pre-Exp 4: the effect of jump timing and fractal images.}
\begin{tabular}{@{}c@{~}|@{~}r@{~}r@{~}r@{~}r@{~}r@{~}r@{~}r@{~}r@{~}r@{~}r@{}} 
    \hline
    Jump timing & \#1 & \#2 & \#3 & \#4 & \#5 & \#6 & \#7 & \#8 & \#9 & \#10 \\
    \hline \hline
    1,000  &6.39&6.36&6.81&6.55&6.85&7.34&7.42&6.31&6.75&7.41\\
    2,000  &8.80&8.25&8.09&8.96&7.96&8.98&9.26&7.96&8.88&8.05\\
    5,000  &9.58&8.59&8.26&9.04&8.75&9.85&8.37&9.30&9.41&9.36\\
    10,000 &10.52&11.13&11.36&11.68&10.32&11.45&10.15&11.36&10.01&11.13\\
    \hline
\end{tabular}
%
\label{tab:Experiment_jump}
\end{table}

\begin{figure}[t]
  \centering
  \includegraphics[width=1.0\linewidth]{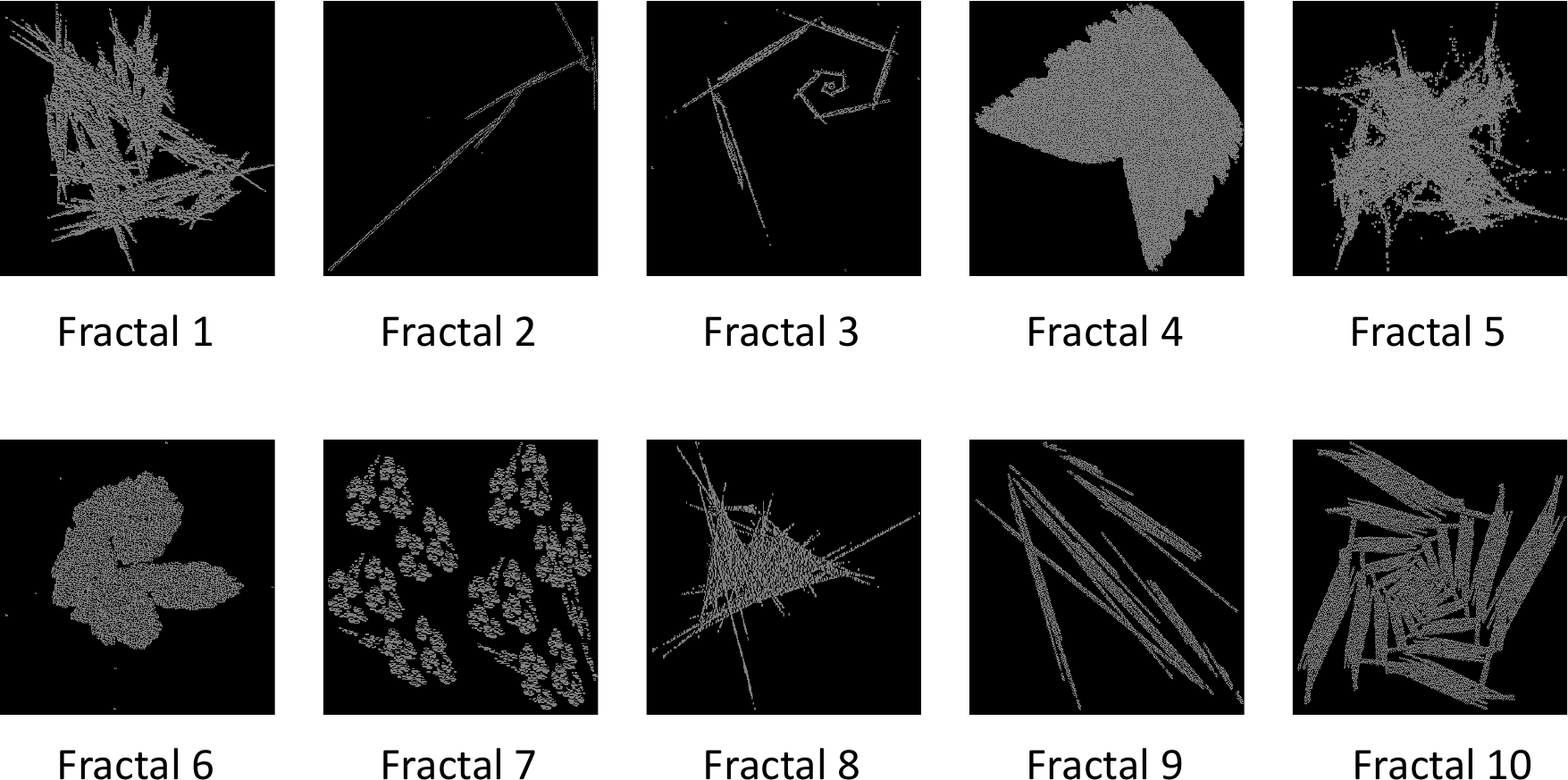}
  \caption{Fractal images used in this experiment.}
  \label{fig:Fractal}
\end{figure}
\begin{figure*}[t]
  \begin{minipage}[b]{0.24\hsize}
    \centering
    \includegraphics[width=4.5cm]{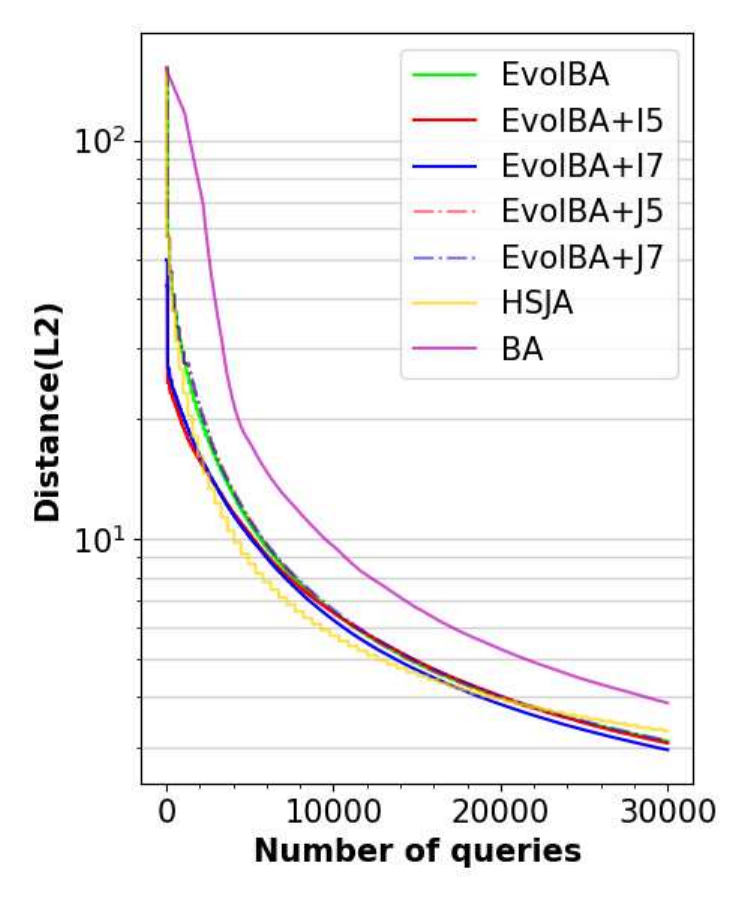}
    \subcaption{VGG19}
  \end{minipage}
  \begin{minipage}[b]{0.24\hsize}
    \centering
    \includegraphics[width=4.5cm]{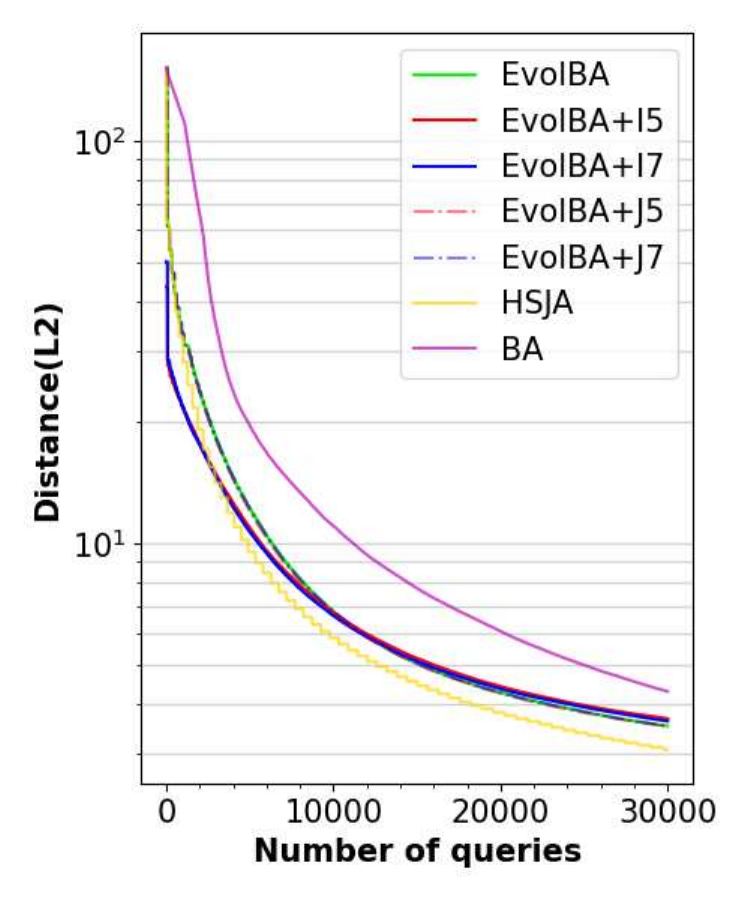}
    \subcaption{ResNet-50}
  \end{minipage}
  \begin{minipage}[b]{0.24\hsize}
    \centering
    \includegraphics[width=4.5cm]{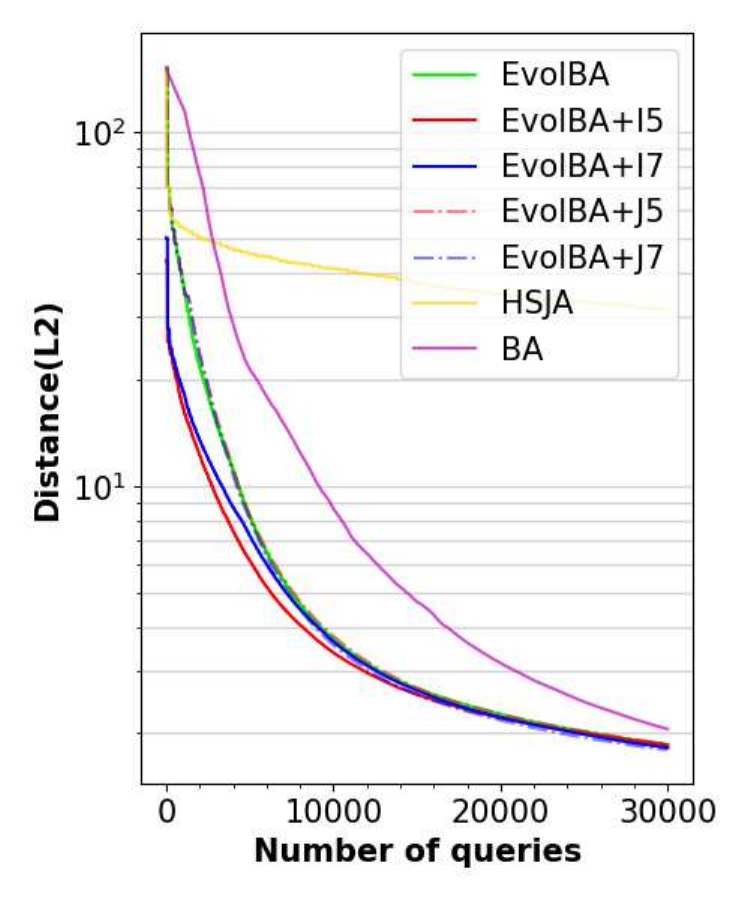}
    \subcaption{Inception-v3}
  \end{minipage}
  \begin{minipage}[b]{0.24\hsize}
   \centering
   \includegraphics[width=4.5cm]{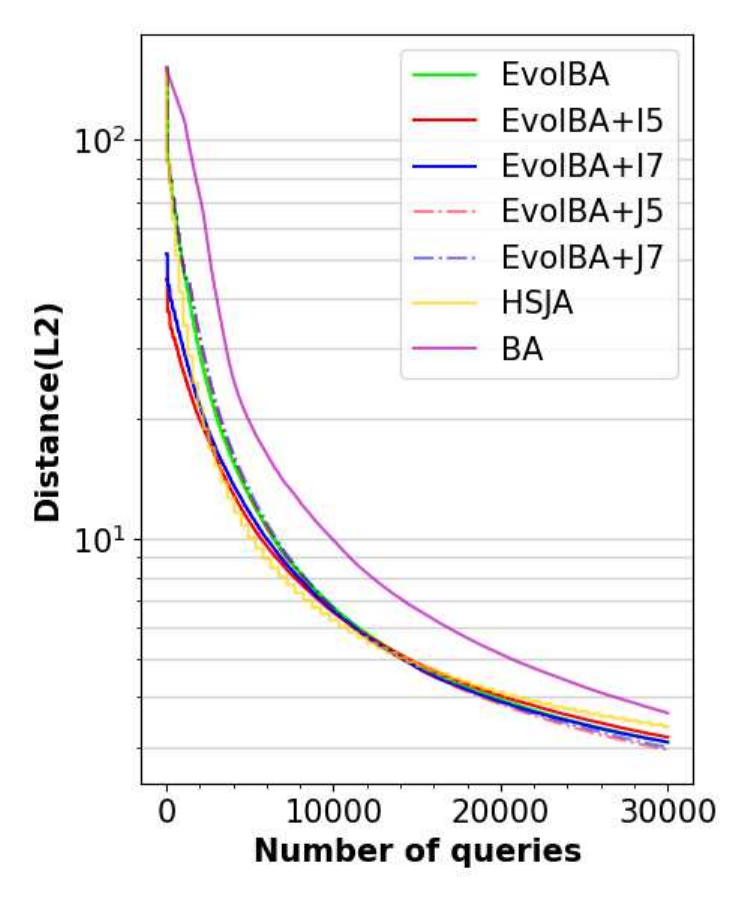}
   \subcaption{ViT}
  \end{minipage}
  \caption{Comparison on L2 distance to the original images averaged over 50 images.}
  \label{fig:result_x5}
\end{figure*}
\begin{table*}[t]
  \centering
    \caption{Comparison on Mean L2 distance at various model queries.}
  \begin{tabular}{@{}c@{~}|@{~}r@{~}r@{~}r@{~}r@{~}|@{~}r@{~}r@{~}r@{~}r@{~}|@{~}r@{~}r@{~}r@{~}r@{~}|@{~}r@{~}r@{~}r@{~}r@{}}
  \hline
  Attack
  & \multicolumn{4}{@{~}c@{~}|@{~}}{VGG19}
  & \multicolumn{4}{@{~}c@{~}|@{~}}{ResNet-50}
  & \multicolumn{4}{@{~}c@{~}|@{~}}{Inception-v3}
  & \multicolumn{4}{@{~}c@{}}{ViT}\\
  \cline{2-17} 
  method&q=2k&q=10k&q=20k&q=30k&q=2k&q=10k&q=20k&q=30k&q=2k&q=10k&q=20k&q=30k&q=2k&q=10k&q=20k&q=30k\\
  \hline \hline 
  EvolBA    &20.30&6.51&3.99&3.09 &23.88&6.78&4.26&3.52&21.54&3.70&2.26&1.85 &28.98&6.71&3.95&3.08\\
  EvolBA+I5 &15.67&6.51&4.02&3.06 &17.77&6.73&4.41&3.67&12.15&3.37&2.20&1.85 &19.78&6.53&4.03&3.17\\
  EvolBA+I7 &16.16&6.24&3.83&2.94 &17.58&6.61&4.35&3.62&13.42&3.62&2.22&1.82 &21.49&6.58&3.90&3.08\\
  EvolBA+J5 &21.30&6.54&3.98&3.07 &23.58&6.72&4.24&3.51&23.55&3.76&2.27&1.86 &31.80&6.62&3.83&2.95\\
  EvolBA+J7 &21.31&6.62&4.02&3.11 &23.88&6.78&4.26&3.52&23.22&3.55&2.17&1.79 &32.10&6.69&3.87&2.99\\
  HSJA      &16.11&5.70&3.96&3.28 &19.21&5.82&3.80&3.07&50.32&41.15&34.87&31.48 &21.35&6.23&4.11&3.37\\
  BA        &75.81&9.57&5.27&3.86 &65.50&11.05&6.05&4.28&75.70&8.62&3.16&2.05 &72.22&9.90&5.13&3.65\\
  \hline
  \end{tabular}
  \label{tab:result_l2_non-hard}
\end{table*}

\section{Preliminary experiments}
\label{sec:pre-exp}

Preliminary experiments were conducted to investigate effective
control parameter settings for EvolBA.
In this experiment, how the performance of the proposed method is
affected by parameters $\sigma$ in Pre-Exp 1, $c_{\mu}$ in Pre-Exp 2, which is a learning
rate of rank-$\mu$ update.
The parameters of Sep-CMA-ES were set to the recommended values given
in Hansen et al.~\cite{hansen2016cma}.
However, for the application to Sep-CMA-ES, we multiplied the learning
rates for rank-one and rank-$\mu$ updates 
$c_1 = \frac{2}{(n+1.3)^2 + \mu_w}$ and
$c_{\mu}=\text{min}(1-c_1,\frac{2(\mu_w-2+\mu_w^{-1})}{(n+2)^2 +
  \mu_w})$ by $(n+2)/3$, respectively.
Also, the number of offspring used to update the vector $\mu$ was set to
the population size, i.e., $\mu=\lambda$.
Under the $\sigma$ and $c_\mu$ derived above, the effect of population
size $\lambda$ was also tested in Pre-Exp 3.

For each experimental condition, we generated AEs on 10 randomly
selected images using the pre-trained VGG19 model as the attack
target.
The optimization
termination condition, i.e., the maximum number of queries to the
model, was set to 30,000 queries. 
In these experiments, to see the changes of Sep-CMA-ES, initial
solution was created by uniform random noise and
the jump operator was not used.
As an evaluation criterion, we focused on $l_2$ distance from an
obtained AE to an original image.

Finally, experiment Pre-Exp 4 was conducted to determine when to use
the jump operator of the proposed method and which fractal image to
use.
Note that this study attempts to apply the jump operator when the
predetermined number of queries is reached. 
Applying the jump depending on the optimization situation is our
future work because it is not easy to clarify the appropriate
conditions.
In this experiment, 8 randomly selected original images and 10 fractal
images selected from FractalDB-1k dataset~\cite{kataoka2020pre} shown
in Fig.~\ref{fig:Fractal} were employed.

Table~\ref{tab:Experiment_sigma} shows the results of Pre-Exp 1, i.e.,
the change in the average amount of perturbation in the generated AEs
when $\sigma$ was varied.
The perturbation decreased in the range from $\sigma=1.0$ to
$\sigma=1.0 \times 10^{-3}$,
and then slightly increased as $\sigma$ was decreased,
reaching around 2.50.
Based on this result, the following experiments adopted the value
$10^{-3}$ as $\sigma$, which showed the smallest perturbation.

Table~\ref{tab:Experiment_cmu} shows the results of Pre-Exp 2, i.e., 
the perturbation amount averaged over 10 images for a change in the
value of the learning rate $c_{\mu}$ of the rank-$\mu$ update.
The perturbation of AE generated was the smallest when $c_\mu \times
10^{-1}$.
This value was used in the following experiments.

Table~\ref{tab:Experiment_lambda} shows the results of Pre-Exp 3,
i.e., the average perturbation amount for varying values of the
population size $\lambda$.
When $\lambda=117$, the perturbation of AE generated was the smallest.
Thus, this value was used in the following experiments.

Table~\ref{tab:Experiment_jump} shows the results of Pre-Exp 4, i.e.,
the average perturbations
for
for different values of the timing using the jump operator and 
for different fractal images.
From the table, the amount of AE perturbation generated when the
number of queries reached 1,000 was the smallest, so we decided to use
this value as the timing for using the jump operator.
In addition, fractal images \#5 and \#7 were better than other images,
then these two images were used in the experiments in Sec.~\ref{sec:experiment}.

\section{Evaluation}
\label{sec:experiment}

\subsection{Experimental setup}

To verify the effectiveness of
the proposed
EvolBA, experiments were
conducted to compare
EvolBA
with
BA~\cite{brendel2017decision} and
HSJA~\cite{chen2020hopskipjumpattack}.
The dataset was created by randomly selecting 50 images, ensuring no
overlap of class labels.
The size of the original images was set to $224 \times 224$ pixels to
match the input size of the classifier used.
The total number of variables was $150,528$.

This experiment adopts four pre-trained DNN models,
VGG19~\cite{simonyan2014very}, ResNet-50~\cite{he2016deep},
Inception-v3~\cite{szegedy2016rethinking}, and
ViT~\cite{dosovitskiy2020image}.

This experiment focused on 
untargeted attacks,
and HSJA used random
noises as initial solutions.
Five types of the proposed method were conducted: the proposed method
that did not use initialization and jump operator (EvolBA), the
proposed method that used initialization but did not use jump operator
(EvolBA+I), and the proposed method that did not use initialization
but used jump operator (EvolBA+J).
Furthermore, the latter two methods using the operators were
distinguished by the fractal image they use: EvolBA+I5, EvlBA+I7,
EvolBA+J5, and EvolBA+J7.
Parameters of Sep-CMA-ES were configured based on Sec.~\ref{sec:pre-exp}.
The number of iterations of the binary search used in the proposed method was set to 26 as in HSJA. 
The penalty value $c_{pen}$ was set to $c_{pen}=1,000$.
The upper limit of the number of mean vector update iterations asset
to $\tau=3$.
The cutoff frequency $r$ was set to 25 for the initial solution
generation and $r=50$ for the jump operator.
The termination condition, of the optimization, i.e., the upper limit
of the number of queries to DNN, was set to 30,000 queries with
reference to HSJA.
The quality of the obtained AEs was evaluated using the L2 norm of the
perturbation.

\subsection{Results}

Fig.~\ref{fig:result_x5} and Table~\ref{tab:result_l2_non-hard} show the
transitions of AE perturbation amount obtained by each method for each
target DNN model.
Comparing the proposed method with BA, for all four DNN models, the
proposed method successfully found AEs with smaller perturbations than BA.
Comparing the proposed method with HSJA, HSJA found AEs with smaller
perturbations in ResNet-50; however,
all variations of EvolBA in VGG19, Inception-v3 and ViT
succeeded in
found AEs with less perturbations than HSJA.
In VGG19 and ViT, it can be seen that EvolBA and its variants were
successful in continuously eliminating the perturbation even after
20,000 queries, while HSJA excelled in the early optimization stages.
These results show that the proposed method clearly outperforms BA,
and can find less perturbed AEs in the models that HSJA shows
limited effectiveness.

Next, focusing on the operators in the proposed method, EvolBA+I5 or
EvolBA+I7, which uses fractal images for initial solution generation,
were superior in the four models.
The graphs of EvolBA+I5 and EvolBA+I7 in Fig~\ref{fig:result_x5} show that
the perturbations were small at the initial solution, indicating the
high effectiveness of using fractal features.
For the jump operator, it has been verified that performance can be
enhanced over EvolBA without the jump operator when applied with a
suitable fractal image for models and input images, such as EvolBA+J7
for Inception-v3 and EvolBA+J5 for ViT.
Clarifying appropriate conditions for jumping and the method of selecting
suitable fractal images are future issues.

\begin{figure}[t]
\centerline{\includegraphics[width=9.0cm]{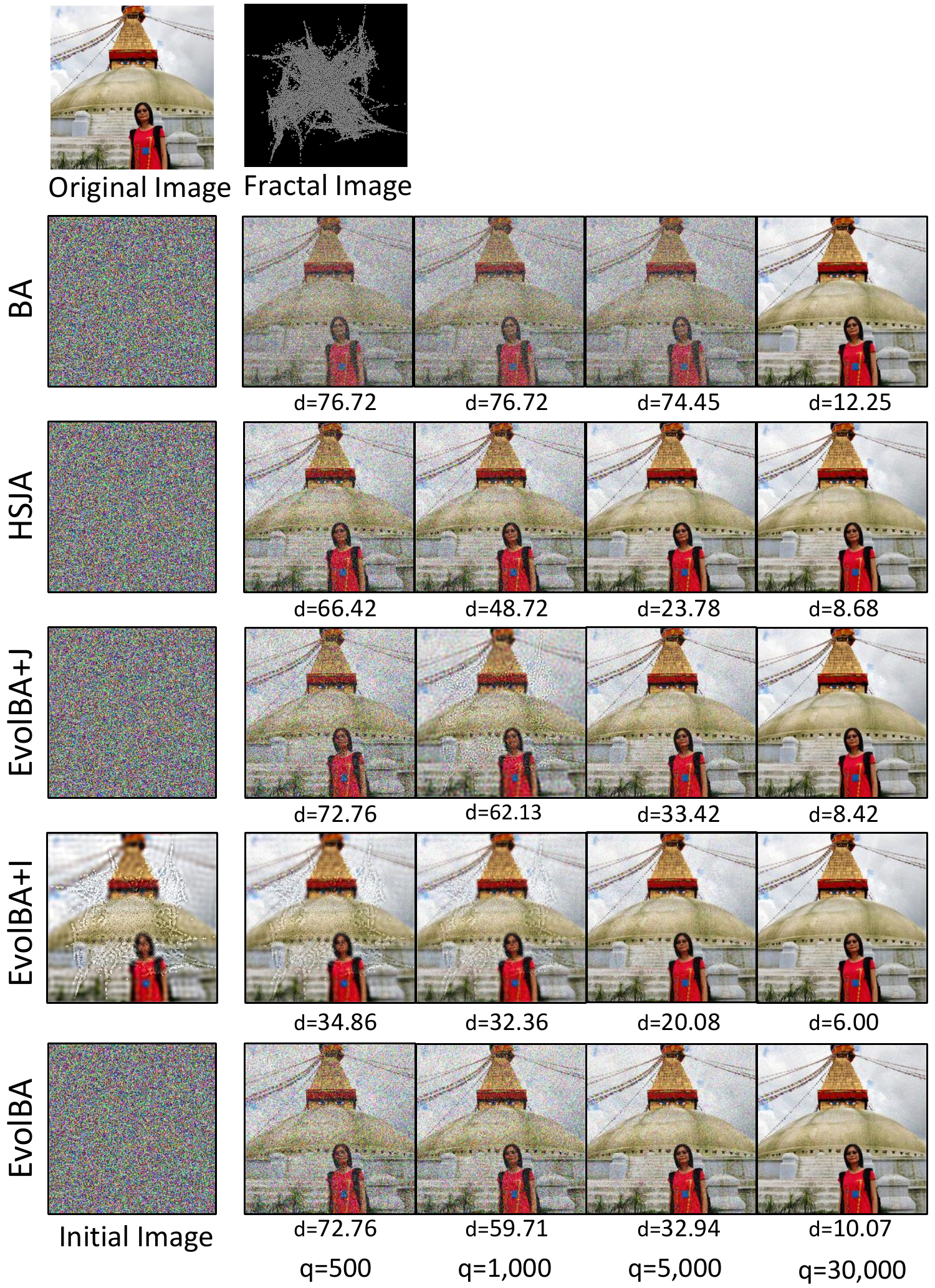}}
\caption{Comparison of attacks on VGG19.}
\label{fig:AE_fig}
\end{figure}

Fig.~\ref{fig:AE_fig} shows examples of the process of generating AE
by each method.
It can be seen that the fractal features are strongly included in the
initial solution of EvolBA+I5 and become less visible as the
optimization proceeds.

\begin{figure}[t]
  \centerline{\includegraphics[width=9.0cm]{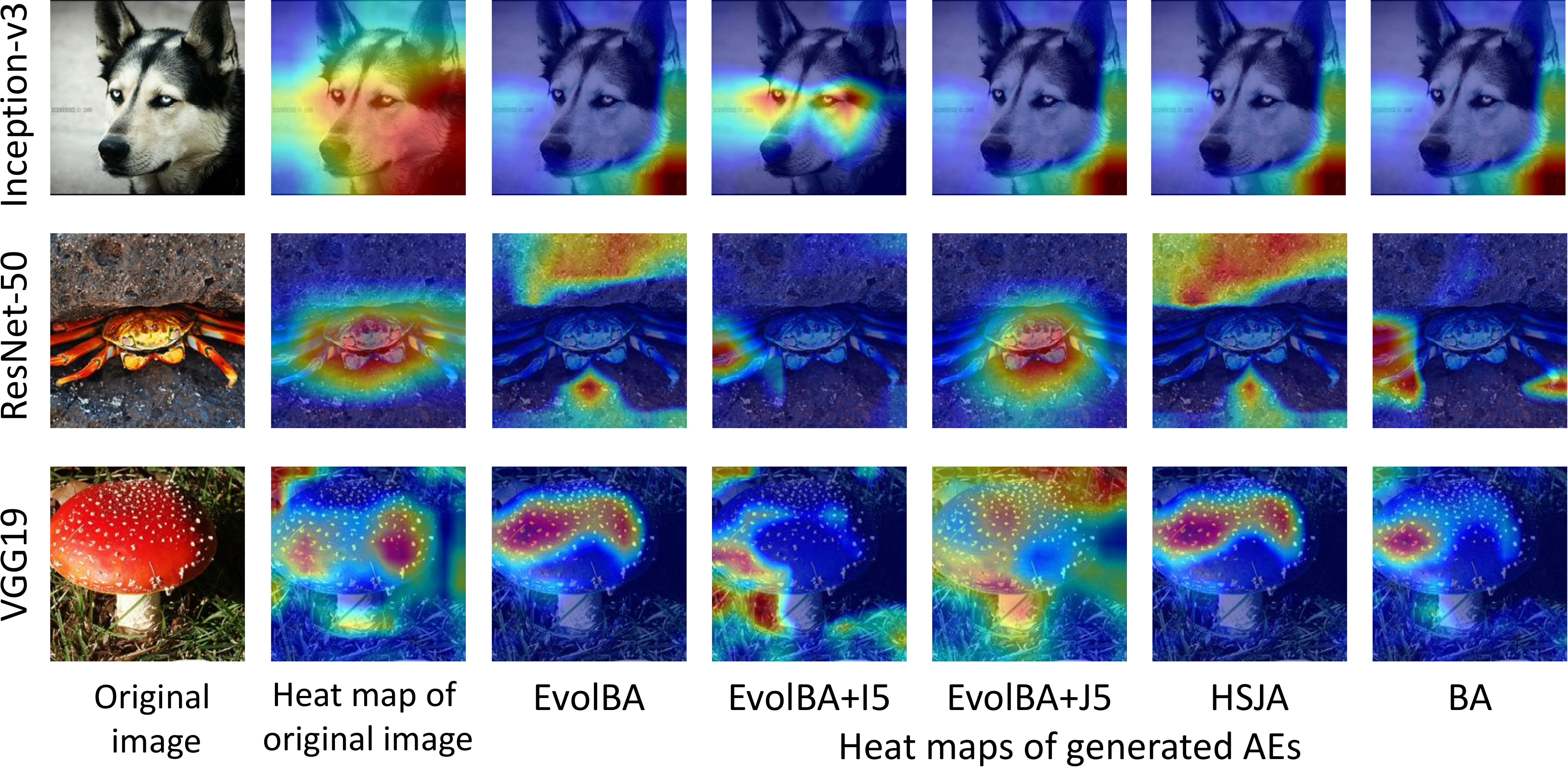}}  
\caption{Class activation maps on generated AEs.}
  \label{fig:AE_gradcam}
\end{figure}

Fig.~\ref{fig:AE_gradcam} shows example class activation maps (CAMs) on AEs
obtained by Grad-CAM.
The fact that the CAMs of EvolBA+I5 were different from others shows
the importance of the initial solution generation.

%

\section{Conclusion}

This paper proposes EvolBA, a method for adversarial attacks under
the HL-BB condition.
The proposed EvolBA generates AEs using CMA-ES, and this paper proposes two 
operators using fractal images that can be used in EvolBA. 
The experimental results showed that EvolBA could stably generate AEs with less
perturbation in images and models compared to BA and HSJA.
We also confirmed that the initial solution generation operator using
fractal features was highly effective.

In future studies, we aim to develop an effective fractal image selection
method and  to search for appropriate jumping conditions.


\section*{Acknowledgment}

This work was supported by JSPS KAKENHI Grant Number JP 22K12196.

\bibliographystyle{junsrt}
\bibliography{reference}

\end{document}